%% file: paper.tex
\documentclass[submission,copyright,creativecommons]{eptcs}
\usepackage{changebar}
\providecommand{\event}{THEdu 2011} 
\usepackage{breakurl}             
\usepackage{enumerate}
\usepackage{subfigure}
\usepackage{xspace}
\usepackage{paralist}
\usepackage{isabelle,isabellesym}
\usepackage{amsmath}
\usepackage{mathpartir}
\usepackage{booktabs}
\usepackage{paralist}
\usepackage[hide]{ed}
\usepackage{tabularx}
\usepackage{amssymb}

\usepackage{tikz}
\usetikzlibrary{arrows,chains,matrix,positioning,scopes}
\usetikzlibrary{positioning,shapes,fit}

\newcommand{\ANDES}{Andes\xspace}
\newcommand{\APROS}{AProS\xspace}
\newcommand{\PARADOX}{Paradox\xspace}
\newcommand{\MACE}{Mace\xspace}
\newcommand{\EPGY}{Epgy\xspace}
\newcommand{\MAPLE}{Maple\xspace}
\newcommand{\OTTER}{Otter\xspace}
\newcommand{\TUTCH}{Tutch\xspace}
\newcommand{\OMEGA}{\textsl{$\Omega$mega}\xspace}
\newcommand{\woz}{Wizard-of-Oz\xspace}

\newcommand{\seq}{\vdash}

\tikzstyle{myboxgray} = [draw=gray!50, fill=gray!20,  thin, rectangle, rounded corners, inner xsep=2pt, inner ysep=2pt]
\newcommand{\highlight}[1]{\tikz[baseline]{ \node[myboxgray,anchor=base] {$#1$}; }}

\newcommand{\nonterminal}[1]{{\it #1}}
\newcommand{\terminal}[1]{{\bf #1}}

\newcommand{\plqed}{{\bf qed}\;}

\newcommand{\plcases}{{\bf cases}\;}
\newcommand{\plsubgoals}{{\bf subgoals}\;}
\newcommand{\plsubgoal}{{\bf subgoal}\;}
\newcommand{\plby}{{\bf by}\;}
\newcommand{\plfrom}{{\bf from}\;}
\newcommand{\plset}{{\bf set}\;}
\newcommand{\pltrivial}{{\bf trivial}\;}

\newcommand{\plassume}{{\bf assume}\;}
\newcommand{\plthus}{{\bf thus}\;}

\newcommand{\plusing}{{\bf using}\;}

\title{Towards an Intelligent Tutor for Mathematical Proofs}
\author{Serge Autexier \qquad\qquad Dominik Dietrich
\institute{German Research Center for Artificial\\ Intelligence (DFKI), Bremen, Germany}
\email{\quad Serge.Autexier@dfki.de \quad\qquad Dominik.Dietrich@dfki.de}
\and Marvin Schiller
\institute{Brunel University, London, UK}
\email{\quad\quad Marvin.Schiller@brunel.ac.uk}
}
\def\authorrunning{Autexier, Dietrich, Schiller}
\def\titlerunning{Towards an Intelligent Tutor for Mathematical Proofs}
\begin{document}
\nochangebars
\maketitle

\begin{abstract}
Computer-supported learning is an increasingly important form of study
since it allows for independent learning and individualized
instruction. 
In this paper, we discuss a novel approach to developing
an intelligent tutoring system for teaching textbook-style
mathematical proofs. 
We characterize the particularities of the domain and discuss common ITS design models.  Our approach is motivated by phenomena found in a corpus of tutorial dialogs that were collected in a \woz experiment.  We show how an intelligent tutor for textbook-style mathematical proofs can be built on top of an adapted assertion-level proof assistant by reusing representations and proof search strategies originally developed for automated and interactive theorem proving.
%
\cbstart The resulting prototype was successfully evaluated on a corpus of
tutorial dialogs and yields good results. \cbend
\end{abstract}

\section{Introduction}
Computer-supported learning is an increasingly important form of study
since it allows for independent learning and individualized
instruction and has resulted in many tutoring systems for different
domains. Mathematics is a key discipline in education and today, there
exist strong systems to teach and tutor specific mathematical skills,
such as mathematical computations, problem solving and geometry (see
for instance,
\cite{Activemath-01-a,KoedingerAndAleven2007,DBLP:conf/aied/MatsudaV05,conf/its/HeffernanC04,conf/lpar/Beeson92,conf/its/NicaudBH02,GeoGebra,DBLP:journals/jar/BillingsleyR07}
to name a few). However, teaching and tutoring support on the ability
of how to do proofs is underdeveloped in state of the art e-learning
systems. Notable exceptions are the tutoring systems for geometrical 
proofs~\cite{DBLP:conf/aied/2005,koedinger1993,trgalova2009}, as well
as for pure formal logic proofs (such as the CMU proof
tutor~\cite{SiegScheines94}, the NovaNet Proof Tutorial
\cite{BarnesAndStamper2010} or Proofweb~\cite{Kaliszyk07}). \cbstart The
overall goal of the work presented in this paper is to provide
e-learning support for classical textbook-style proofs, which is not
fostered by the above approaches. \cbend

Following Van Lehn (see \cite{DBLP:journals/aiedu/VanLehn06}),
intelligent tutoring systems (ITSs) can be characterized as having
both an outer loop and an inner loop. The outer loop selects a
relevant task (exercise) for the student to complete. The inner loop
iterates over individual problem-solving steps, evaluates the steps
and provides feedback to the student. 
Typically, this is achieved by providing the following services:
\begin{inparaenum}[(S1)]
\item \label{service:minimal feedback} minimal feedback on a step, 
\item \label{service:error feedback} error specific feedback on an
  incorrect step, 
\item \label{service:hints} hints on the next step, 
\item \label{service:knowledge assessment} assessment of
  knowledge, and 
\item \label{service:solution review} review of the solution.
\end{inparaenum}

\cbstart In this paper we focus on the
inner loop and report on how we adapted the proof assistant system
\OMEGA to realize the services ~(S\ref{service:minimal feedback}),
(S\ref{service:error feedback}), (S\ref{service:hints}) and a
bit of ~(S\ref{service:solution review}). \cbend
The services are provided by two components, one for \emph{step
  analysis}, and one for \emph{step generation}. The step analyzer
produces minimal feedback on a student's proof step, such as whether
the step is correct or not, together with additional information that
might be used to provide more sophisticated feedback, e.g., if a
specific error type could be extracted. The step generator returns
information about a step the student should do next, e.g., when a hint
is requested by the student.
\ednote{DD: Ich finde den finally satz komisch hier, streichen? Finally, as an outlook we
discuss which knowledge and techniques are available in the proof
assistant system after and during a tutoring session to facilitate the review the
solution~(S\ref{service:solution review}) and to assess the knowledge
~(S\ref{service:knowledge assessment}) of the student.\\AS: gestrichen.}

Individual aspects of this work have been published before and
describe different stages of the development of the different
components.  For instance, parts of the step analyser described
in~\cite{marvinAKA2010} have never been described in connection with the other
parts of the final step analyzer or the final hint generation module~\cite{dietrich2011}.
The contribution of this paper is to provide a coherent overview of
all components from the initial requirements analysis, via the design
phase to their final implementation and evaluation.

\cbstart
The paper is organized as follows: Section~\ref{sec:requirement
  analysis} discusses the aspects of the tutoring problem by analyzing
the teaching domain and reports on a Wizard-of-Oz experiment 
\cbstart[4pt]%
which was part of the design phase of our ITS system. The collected 
data in the form of tutorial dialogs represents 
 a kind of ``gold standard'' which we try to \cbend
approximate. Informed by the analysis of the collected data, we
briefly review the state of the art \cbstart[4pt]
in 
designing tutoring systems and \cbend
derive the architecture of our tutoring system. The methods and tools to analyze
student's proof steps are presented in Section~\ref{sec:step
  analysis}.  Section~\ref{sec:hint generation} presents how hints
with an increasing degree of explicitness are generated from the
domain specific proof strategies specified by the domain expert.  The
intelligent tutor services have been evaluated on the corpus of tutorial
dialogs about proofs 
\cbstart[4pt]
as presented in 
\cbend
Section~\ref{sec:evaluation}. Section~\ref{sec:future} presents an
outlook of further qualitative service improvements that are within
grasp. We review related work in Section~\ref{sec:related work} and
conclude in Section~\ref{sec:conclusion}. \cbend

\section{Requirements Analysis and Functional Specification}
\label{sec:requirement analysis} 

The teaching goal in our domain is that students develop the skill to
conduct and author mathematical proofs in textbook style. Our target
students are in the final high-school years or undergraduate
university students. Therefore, the assumptions on the students are
that they have a knowledge of specific mathematics domains, but not
necessarily of mathematical logic.
Textbook-style proofs themselves consist of intermediate proof steps
in a declarative style stating a subgoal or a derived fact. In that
sense they are compatible to formal logic proofs. In contrast, the
justifications in textbook style proofs are typically not exclusively
references to basic formal logic rules. Rather, they are justified by
references to hypotheses, definitions, lemmas or
theorems---collectively called \emph{assertions}---as well as specific
proof strategies (e.g., well-founded induction) or symbolic
computations (e.g., polynomial factorization)---collectively called
\emph{reasoning techniques}.\ednote{DD: added this sentence here, do
  we want to keep it? MS: Yes, it makes sense to me. AS: Yes.} It is
common practice to reduce the size of a proof by leaving out
references to assertions or reasoning techniques if they can easily be
inferred by the reader.

\paragraph{Solution Space.} 
\ednote{DD: what about equivalent reformulations here, e.g., direct
  vs. indirect argumentation of the same proof idea\\AS: Yes, but I
  can't se immediately how to fir it in. Let's do it for the final
  version.}  For a given theorem, there is typically a space of
possible proofs. 
This is for
 a variety of reasons: First, from a mathematical point of view, 
there often exist various distinct proofs for the same proof problem. 
 For example, Ribenboim
gives eleven proofs that there are infinitely many primes (see
\cite{Ribenboim96} for details). Especially, given a set of proofs it
is difficult to be certain 
 that no further mathematically sensible proof
is possible, i.e., that a given set of proofs contains all
mathematically sensible proofs.

Secondly, even for a specific proof, there are different 
 ways to 
formulate the proof: permutability of proof steps is one reason which
essentially leaves the proof structure invariant. Additionally, a
proof step can be formulated in forward-style by deriving new facts or in
backward-style by introducing new subgoals: the choice of
forward-style vs. backward-style has a more severe impact on how the
proof is structured.  Typically, in textbook-style proofs most 
steps are in forward-style because it 
makes it easier to follow the
proof. However, proof search often happens in backward-style, and the proofs 
are reformulated in forward-style afterwards (see \cite{Solow05} p. 13-15 for a discussion).

A given proof with a specific proof-style and a particular order of
applying assertions is 
considered to be at the lowest level of mathematical proof, 
as all information is
explicitly provided. However, this is often much too detailed and more
high-level versions of the proof are also acceptable or preferred. For instance,
one may allow for larger proof steps (w.r.t. step-size) 
and the omission of those 
justification details that are considered as trivial. The step-size of proof steps
and the detailedness of their justifications directly correlates with
the conciseness of a proof, which in turn is crucial in order to
effectively communicate the idea of a proof. Which high-level proof is
actually still acceptable depends, of course, on the expertise of the
audience; for instance, proofs in introductory mathematical textbooks
are much more detailed than in advanced mathematical textbooks. 
Thus,
thirdly, the detailedness of proofs adds a further dimension along
which a proof may vary.

\paragraph{Proof Analysis Criteria.} When judging a proof there are
local and global aspects: On a local scale, one may judge each proof
step separately. The first and foremost criterion, of course, is 
 whether the proof step is
correct or not. For an incorrect proof step, it is important to know
why it is wrong in order to be able to repair or remove it.  On
the other hand, a correct proof step is not necessarily relevant for a
proof, such as tautological steps or redundant proof steps. Thus, a
second criterion is the relevance of a proof step. Furthermore,
conciseness is important to convey the idea of a proof. On the local
scale of proof steps this boils down to judging if a proof step is of
adequate size and whether the details of its justification are
appropriate. Finally, for textbook proofs it matters whether a
backward-style proof step should preferably be re-formulated into a
forward step or vice versa.
On a global scale, the conciseness of a proof in the sense of whether
the proof is without detours, or if there exist alternative shorter
proofs, 
\cbstart[4pt]
are interesting aspects.
\cbend

\paragraph{Proof Construction.} The goal to write proofs in textbook style
requires 
knowledge of how to find a proof in the first place. Both the construction of 
a proof as well as the ability to come up with a good presentation of
the proof which conveys the proof idea well requires proof strategic
knowledge, as well as knowledge about when which proof strategy is appropriate or
inappropriate. A human teacher tries to develop that knowledge by
training the students on proof problems requiring a specific
strategy. When students get stuck in the proof, trying to help them by
strategic advise is the primary choice instead of showing them the
detailed next steps.

\subsection{\woz Experiment}
\label{sec:woz study}

Our approach to the intelligent tutoring of proofs is informed\ednote{DD: das hoert sich komisch an, finde aber keine gute loesung? MS: ist aber haeufig, google mal ``our approach is informed by''... AS: Fine with me.}
 by experiments to study %
\cbstart[4pt]
the 
\cbend
specific requirements for such a system. 
In particular, we used the \woz paradigm\footnote{\woz 
 experiments~\cite{kelley1984} simulate a complex system via a partial/prototype implementation that is assisted 
by a human expert (the ``wizard''). Such experiments provide valuable 
data to assist the design and to evaluate components of such a system
 in advance of its completion.} to assess the 
requirements for the system's sub-components to fulfill the
 tasks of the inner loop that are specific to proof tutoring, 
as outlined in the previous section.


The data was collected in an experiment where thirty-seven students
interacted with a mock-up of a natural language dialog tutoring system
for mathematical proofs.  The system was simulated via a specific
software environment \cite{C22} and the help of four experienced human
tutors.  We obtained a corpus of tutorial dialogs
\cite{dialogcorpusLREC06} that allowed us to study the actions of
students and tutors related to proof exercises illustrating the
properties of binary relations.  Student input consisted of natural
language text and formulas, to investigate the prospect of
natural-language understanding for mathematics within such a tutoring
system.
In addition to providing feedback to student's actions (proof steps, questions or comments),  
the tutors rated each proof step with respect to correctness, granularity (or
proof step size) and relevance to the current task. 

Figure~\ref{corpus:st25p2} shows a fragment of a tutorial session in
which the student was instructed to prove the theorem $ ( R
\circ S )^{-1} = ( S^{-1} \circ R^{-1} )$, where $R$ and $S$ are
relations, and $\circ$ and $^{-1}$ denote relation composition and relation
inverse, respectively. In the examples, {\bf S} refers to a student turn and {\bf T} to a tutor turn.
\begin{figure}
\centering
\begin{tabularx}{\linewidth}{|X|X|X|} 
  \hline& & \\[-4mm]
  \begin{tabular}[t]{l}
    {\bf S8:} let $(x,y) \in \left(R\circ S\right)^{-1}$\\
    {\bf T9:} correct \\
    {\bf S10:} hence $(y,x) \in (S \circ R)$\\
    {\bf T11:} incorrect \\
  \end{tabular} 
  & 
  \begin{tabular}[t]{l}
    {\bf S8a:}  we consider the subgoals \\
    $\left( R \circ S \right)^{-1} \subset S^{-1}\circ R^{-1}$ \\ 
    and $\left( R \circ S \right)^{-1} \supset S^{-1}\circ R^{-1}$ 
  \end{tabular}
  &
  \begin{tabular}[t]{l}
    {\bf S8b:}  first, we consider the\\
     subgoal $\left( R \circ S \right)^{-1} \subset S^{-1}\circ R^{-1}$
  \end{tabular}
  \\\hline
\end{tabularx}
\\
\caption{Examples illustrating phenomena of the corpus}
\label{corpus:st25p2}
\vspace{-0.3cm}
\end{figure}
The approach taken by the student in the first example on the left of
Figure \ref{corpus:st25p2} is to apply set extensionality and then to
show that the subset relation holds in both directions. The student
begins in utterance {\bf S8} by directly introducing a pair $(x,y)$ in
the set $(R \circ S)^{-1}$. This is rated as correct by the tutor, who
recognizes that the student wants to prove both directions separately
and that the introduction of the pair $(x,y)$ is useful due to the
definition of subset.
The student then states an incorrect formula in {\bf S10}, which the tutor
rates as incorrect.

Two alternative ways that the student could have started the same
exercise, which we will use as running examples in this paper, are
shown on the right in Figure \ref{corpus:st25p2}. In {\bf S8a} the
student explicitly splits the proof into two subgoals with an
application of set extensionality.  In {\bf S8b} the same rule is
applied, but only one of the two resulting proof obligations is
explicitly presented.
\vspace*{1ex}

We analyzed those utterances from the corpus which contain
contributions to the theorem proving task. We were able to identify
\cbstart five general phenomena which \cbend must be accounted for in
order to correctly verify (or reject) the proof steps that students
perform and to maintain correct consistent representations of the
proofs they are building. \cbstart These phenomena show that verification in
this scenario is not simply a matter of logical correctness, but must
also take into account the proof context, for instance. \cbend

\paragraph{Underspecification.} Some subset of the complete description of a
proof step is often left unstated. Utterance {\bf S8} is an example of
a number of different types of this underspecification which appear
throughout the corpus. \cbstart The proof step in {\bf S8} includes the
application of set extensionality, but the rule and its parameter are
not stated explicitly. The student also does not specify that of the
two subgoals introduced by set extensionality, he is now proving one
particular subset direction, nor does he specify the number of steps
needed to reach this proof state. \cbend Part of the task of analyzing such
steps is to instantiate the missing information so that the formal
proof object is complete.

\paragraph{Incomplete Information.} Proof steps can, in addition to
issues of underspecification, be missing information which is necessary for
their verification by formal means. 
\cbstart For instance, utterance {\bf S8b} is a correct contribution to the proof, but the second subgoal is not stated. This second
subgoal is however necessary to verify that proving the subset relation is part
of justifying the equality of the sets, since one subgoal alone does not imply
the set equality which is to be shown. \cbend

\paragraph{Ambiguity.} Ambiguity pervades all levels of the analysis of
the natural language and mathematical expressions that students
use. Even in fully specified proof steps an element of ambiguity can
remain. For example in any proof step which follows {\bf S8a}, we
cannot know which subgoal the student has decided to work on.
Also, when students state formulas without indicating a proof step
type, such as ``hence'' or ``subgoal'', it is not clear whether the
formula is a newly derived fact or a newly introduced
subgoal. Again, this type of ambiguity can only be resolved in the
context of the current proof, and when resolution is not possible, the
ambiguity must be maintained by the system.

\paragraph{Proof Step Granularity.}

As outlined above, proofs can 
 generally be constructed 
at various levels of detail. \ednote{DD: Marvin, hast du da nicht ein schoenes ergebnis aus deiner dipl. arbeit, das man hier einfuegen koennte? MS: Leider nein -- ich kenne nur vom Hoeresagen, dass elegante Loesungen fuer vormals sehr lange Beweise gefunden wurden, die  auf einen Bierdeckel passen -- aber ich kann mir die Namen nicht merken, weil ich die Theorien gar nicht kenne.} In a tutorial setting, however, 
the tutor needs to ascertain that the 
student develops the proof at an acceptable pace. For this task, 
 classical reasoners are of little help, 
since they usually provide large proof 
 objects based on some particular logical calculus, such as
 resolution, that operate at a different step size (granularity) 
 than typical mathematical practice. 
Generally, typical proof steps may represent several steps in a more formal 
representation, such as the natural deduction calculus or proofs 
at the assertion level. This is even the case for proofs at the beginner
level (cf.~\cite{C25}).



 In the \woz  
studies, the tutors were found to react to deviations in step size, 
e.g.~in the dialog fragment in Figure~\ref{granularityexample}, 
where the student 
is skipping a sub-step the tutor expects to see. Generally, 
whether a proof is of acceptable step size (granularity) depends 
on the student's knowledge (which we represent via a simple 
 overlay student model) 
and other factors.  

\begin{figure} 
 \fbox{\begin{tabular}{p{0.04\textwidth}p{0.89\textwidth}}
    {\bf T:} & [Show] $(R \cup S) \circ T=(R \circ T) \cup (S \circ T)$\\
    {\bf S8:} & \begin{minipage}[t]{0.9\textwidth} 
          $(a,b)  \in (R \cup S)$, if $(a,b)  \in R$ or
          $(a,b)  \in  S$
         \end{minipage} \\
     {\bf T:} &  \begin{minipage}[t]{0.9\textwidth} 
           Correct. \ \ 
 \begin{tabular}{|l|l|l|} 
 \hline correct & appropriate & relevant \\ \hline \end{tabular} 
         \end{minipage} \\
      {\bf S9:} & \begin{minipage}[t]{0.9\textwidth} 
           $\exists x$, such that $(a,x)  \in  (R \cup S)$ and $(x,b)  \in 
          T$
         \end{minipage} \\
       {\bf T:} &  \begin{minipage}[t]{0.9\textwidth} 
          What does this follow from? \ \ 
 \begin{tabular}{|l|l|l|} 
 \hline correct & too coarse-grained & relevant \\ \hline \end{tabular} 
         \end{minipage} \\
       {\bf S10:} & \begin{minipage}[t]{0.9\textwidth} 
           This follows from the definition of relation product for binary
          relations: $(a,x)  \in R$ and $(x,b) \in S$, hence 
           $R \circ S$
         \end{minipage} \\
       {\bf T:} &  \begin{minipage}[t]{0.9\textwidth} 
           More concretely: it follows from $(a,b) \in (R \cup S) \circ T$
          \ \ 
 \begin{tabular}{|l|l|l|} 
 \hline correct & appropriate & relevant \\ \hline \end{tabular} 
 
         \end{minipage} \\
\end{tabular}}

\caption{Dialog fragment illustrating tutor's intervention for inappropriate
  granularity of {\bf S9}}
\label{granularityexample}

\end{figure}


\paragraph{Relevance of Proof Steps.}

The tutors in the experiments indicated when they believed 
that steps suggested by the student were not goal-directed. 
One problem that we address in this work is that  
the student may introduce hypotheses (e.g.~``let $(x,y) \in (R \circ S)^{-1}$'' 
in Figure~\ref{corpus:st25p2}), which 
may or may not be useful with respect to the current proof goal. 
We refer to this form of assessment as \emph{relevance checking.}

\subsection{Domain Modelling \& Teaching Strategies\ednote{To all: I tried to combine the previous titles a little bit -- please change/revert if you don't like it (M)}}
\label{subsec:tutoring}

ITS are designed to be effective -- i.e., to lead to increased knowledge and skill via engaging the student 
with the system. There are three main approaches in the literature on how to build an
ITS: Model tracing tutors, constraint based tutors, and example
tracing tutors:

\noindent\textbf{Model Tracing Tutors (MTTs)}, such as the Andes
physics tutor (see \cite{DBLP:journals/aiedu/VanLehnLSSSTTWW05}),
contain a \emph{cognitive model} of the domain that the tutor uses to
``trace'' the student's input, i.e., to infer the process by which a
student arrived at a solution. MTTs are based on the ACT-R theory of
skill knowledge \cite{actrtheory} that assumes that problem solving
skills can be modeled by a set of \emph{production rules}. Given a
student input, a \emph{model tracer} then uses these rules to find a
\emph{trace}, i.e., a sequence of rule applications that derive the
student's input. If such a trace can be found, the student is assumed
to have used the same reasoning as encoded in the rules to arrive at his
input and the step is reported to be correct
(cf.~(S\ref{service:minimal feedback})). Thus, the tutor can use the
trace to analyze the cognitive process of the student. Alternative
solutions are supported by providing rules that capture alternative
solution approaches.

To be able to also trace common student errors, a MTT typically
provides a set of \emph{buggy rules} (see \cite{buggyrules}) that
model incorrect reasoning. If a trace contains one or several buggy
rules, the step is assumed to be incorrect and error specific feedback
can be given to the student (cf.~(S\ref{service:error feedback})).



In practice, it turns out that a model tracer will not always be able
to trace all student inputs, for example, if a solution cannot be
found due to the complexity of the search space, or because a faulty
step is not captured by a buggy rule. In this case, it is either
assumed that the student step is wrong or an undefined answer is
returned by the analysis component. 

MTTs can offer strategic and context sensitive problem-solving hints
on demand by computing a solution for the current proof state using
the expert module (cf.~(S\ref{service:hints})). By analyzing this
solution, hint sequences can be computed that contain increasingly
more information about the next step to be performed. The dynamic
generation of the solution guarantees that the hint will be tailored
to the specific situation in which the student got stuck.

\medskip
{\bf Constraint Based Tutors (CBTs)}, such as the SQL tutor (see
\cite{DBLP:conf/its/SuraweeraM02}), are based on Ohlsson's theory of
learning from performance errors (see \cite{6551624}) and use
\emph{constraints} to describe abstract features of correct
solutions. There are two fundamental assumptions:
\begin{enumerate}[(i)]
\item Correct solutions are similar to each other in that they satisfy all
the general principles of the domain. 
\item Diagnostic information is not contained in the sequence of
actions leading to the problem state, but solely in the problem state
itself.
\end{enumerate}
Constraints describe equivalent student states and consist of three
components: a \emph{relevance condition}, a \emph{satisfaction
  condition}, and a \emph{feedback message}. The relevance condition
describes the abstract properties of the class of solution states that
is represented by the constraint. The satisfaction condition contains
additional checks a state of this class must satisfy in order to be
correct. The feedback message contains the feedback that is given to
the student when the satisfaction condition is not satisfied, i.e.,
when an error is detected (cf.~(S\ref{service:error
  feedback})). If the student enters a situation where the tutor has
no knowledge of, i.e., no relevance condition evaluates to true, the
CBM remains silent (cf.~S(\ref{service:minimal feedback})).
Typically, CBTs provide two kinds of constraints: syntactic
constraints and semantic constraints. Syntactic constraints check
whether the input is well-formed, whereas semantic constraints compare
the input with an optimal solution provided by the tutor. Semantic
constraints also check for alternative solutions by capturing
alternative subexpressions of a solution.

 

As CBTs are not equipped with an expert system that solves problems,
they cannot automatically complete solutions for a given problem
state. Therefore, hints can only be given by comparing the current
solution state with an ideal solution, trying to detect missing
features. This entails the risk that the hints that are given are
overly general or misleading if the student follows a solution
different to the ideal solution that is given by the tutor
(cf.~(S\ref{service:hints})).

\medskip
{\bf Example Tracing Tutors (ETTs)}, such as the stoichiometry tutor
(see \cite{mclaren2008}), interpret a student's solution step with
respect to a predefined \emph{solution graph} that represents a
generalized solution, which is often also called \emph{behavior graph}
(cf.~\cite{NewellSimon72}). A behavior graph is a directed, acyclic
graph, whose nodes represent problem solving states and whose edges
represent problem solving actions. Several outgoing edges represent
different ways of solving the problem represented by the state
corresponding to the node. Misconceptions and common errors can be
included within the graph using so-called \emph{failure links} that
indicate typical failures. This way, ETTs can give specific feedback
to both correct and incorrect steps (cf.~(S\ref{service:minimal
  feedback}, S\ref{service:error feedback})).

Initially, the current student's state is the root node of the graph,
which is the only node that is marked as visited. Given a student's
input, the input is matched against all outgoing edges of the current
node. If the input matches a regular link leading to a yet unvisited
node, the step is classified to be correct, if it matches a failure
link or no link at all, it is classified to be incorrect. If a
step matches multiple regular links, all successor nodes represent 
\emph{possible interpretations} of the student's step, which are then
maintained in parallel. Behavior graphs can be extended to
\emph{generalized behavior graphs}, e.g., by defining groups of
unordered steps (so the student can change the order of the steps), or
by generalizing the matching condition for a link.

Behavior graphs are also used to provide hints as to what a student
might do next (cf.~(S\ref{service:hints})). This is done by
identifying an unvisited link in the behavior graph, and then
displaying a hint message associated with that link. ETT have the
advantage that they do not require to model domain knowledge in form of
production rules or constraints and are therefore considerably cheaper
to develop. However, they work only for solutions that were foreseen
by the author of the exercise.

\begin{table}[t]
\centering
\begin{tabular}{lp{0.22\textwidth}p{0.22\textwidth}p{0.22\textwidth}}
\toprule
  Property & MTT & CBT & EBT \\
\midrule
feedback correct step & yes & no & yes\\
error specific feedback & buggy rules & satisfaction condition & error paths\\
hints & context sensitive and strategic & only missing elements & strategic\\
authoring & production rules and strategies & constraints and ideal solution & all solution paths\\
runtime costs & high & low & low\\
step size diagnosis & yes & no & half\\
alternative solutions & yes & yes & yes\\
diagnosis no match & error & correct & error \\
theory & ACT-R & Ohlsson's theory of learning & - \\
\bottomrule
\end{tabular}
\caption{Comparison of MTT, CBT, and EBT}
\label{tab:comparisontutormodels}
\end{table}

\paragraph{Summary.} Table \ref{tab:comparisontutormodels} summarizes the properties of the
different approaches to design an ITS.

\subsection{Functional Specification for the Proof Tutoring System}
\label{sec:function specification}

For our \woz experiments we
deliberately did not impose any restrictions on the language used to
write proof steps, and the input varied from pure formulas to pure
natural language and all forms of mixed natural language and formulas
(see the example fragment shown in
Figure~\ref{corpus:st25p2}). Processing that input poses challenging
problems for natural language understanding. In order to have a clear
separation of concern, we devised a clear, formal interface language
for the kernel module, which serves as target for the natural language
analysis component(s) that still need to be developed
(see~\cite{springerlink:10.1007/978-3-540-89408-7_12} for recent work on that
topic). This also has the advantage that different natural language
analysis components for different languages can be used on top of
the kernel module.

\cbstart Our interface language for the kernel module is a declarative
proof language (see for
example~\cite{Microsoft_threetactic,DBLP:conf/tphol/Wenzel99,AF-05-a,DBLP:conf/types/Corbineau07,UCAM-CL-TR-416})
that has been modified to support the elision of information that is
typically required to facilitate the verification process. This is
because declarative proofs that can be processed by current proof
assistants are usually much more detailed than corresponding textbook
proofs that we want to teach. \cbend For example, we do not enforce the
student to give justification hints or restrict the student to a
specific granularity. By allowing arbitrarily large gaps between the
commands, one arrives at the notion of a \emph{proof
  plan}~\cite{DBLP:journals/entcs/DennisJP06} or \emph{proof
  sketch}~\cite{Wiedijk02formalproof}. Following van Lehn's
requirement that an ITS should allow a student to stepwise construct a
solution, we obtain as individual building blocks single declarative
proof commands. This approach has the following properties:
\begin{enumerate}
  \item Proof commands are the primary solution steps a student can enter.
  \item The sum of all proof commands gives a complete solution in the style of a textbook proof.
  \item A proof command is justified in the context of the previously given steps.
  \item Proof steps might be partial, incomplete or underspecified.
\end{enumerate}

\begin{figure}[t]
\fbox{
\begin{minipage}{.97\textwidth}
\begin{minipage}[t]{8cm}
\begin{tabbing}
  \nonterminal{proof} ~~~ \= ::= \terminal{proof} \nonterminal{steps} \terminal{qed} \\
  \nonterminal{steps} \> ::= (\nonterminal{ostep};\nonterminal{steps}) $|$ \nonterminal{cstep} \\
  \nonterminal{ostep} \> ::= \nonterminal{set} $|$ \nonterminal{assume} $|$ \nonterminal{fact} $|$ \nonterminal{goal} \\
  \nonterminal{cstep} \> ::= \nonterminal{trivial} $|$ \nonterminal{goals} $|$ \nonterminal{cases} $|$ $\mathbb{\epsilon}$ $\quad$\\
  \nonterminal{by} \> ::= \plby \nonterminal{name}? $|$ \nonterminal{proof} \\
  \nonterminal{from} \> ::= \plfrom (\nonterminal{label} (, \nonterminal{label})$^*$)? \\
  \nonterminal{sform} \> ::= \nonterminal{form} $|$ \terminal{.} \nonterminal{binop} \nonterminal{form} 
\end{tabbing}
\end{minipage}
\begin{minipage}[t]{6cm}
\begin{tabbing}
  \nonterminal{assume} \= ::= \=\plassume \nonterminal{form}+ \nonterminal{from} \nonterminal{steps} \plthus \nonterminal{form}\\
  \nonterminal{fact} \> ::= \nonterminal{sform} $|$ \nonterminal{by} \nonterminal{from} \\
  \nonterminal{goals} \> ::= \plsubgoals (\nonterminal{goal})$^+$ \nonterminal{by} \\
  \nonterminal{cases} \> ::= \plcases (\nonterminal{form} \{ \nonterminal{proof} \})$^+$ \nonterminal{by} \nonterminal{from} \\
  \nonterminal{goal} \> ::= \plsubgoal \nonterminal{form} (\plusing \nonterminal{form} ({\bf and} \nonterminal{form})$^+$)? \nonterminal{by}\\  
  \nonterminal{set} \> ::= \plset \nonterminal{var}=\nonterminal{form} (\terminal{,} \nonterminal{var}=\nonterminal{form})$^*$\\
  \nonterminal{trivial}\>::= \pltrivial \nonterminal{by} \nonterminal{from}
\end{tabbing}
\end{minipage}
\end{minipage}}
\caption{\OMEGA proof script language}
\label{tbl:abstractgrammar}
\vspace*{-2mm}
\end{figure}

We use \OMEGA's declarative proof script language presented in
Figure~\ref{tbl:abstractgrammar} (see
also~\cite{DBLP:conf/itp/AutexierD10}) as input language and allow
underspecified proof scripts that are obtained by omitting
``\nonterminal{by}'' and ``\nonterminal{from}'' as well as the
``\plthus \nonterminal{form}'' in \nonterminal{assume}-proof steps. We
also added them as special closing proof steps (\nonterminal{cstep})
at the end of \nonterminal{steps} following an introductory
{\bf assume}. Similarly, we relax the required \plqed at the end of a
{\bf proof}.

To develop the intelligent tutor we follow the MTT approach: first,
the size of the solution space of textbook-style proofs with adaptive
degree of detailedness ruled out the EBT approach. Since the solutions
are proof sketches, the constraints that would have to be formulated
when following the CBT approach would have to be constraints on proof
sketches. The formalism required to formulate such constraints comes
close to what is expressed in the tactic and declarative proof script
language of the proof assistant system. The difference is that
constraints are descriptive while information contained in tactics is
constructive, which makes them attractive to generate (strategic)
hints. Finally, the solution objects, i.e. (sketches of) declarative
proofs, are themselves abstract structurings of parts of the
cognitive model representing how the student solves the problem
(proving the theorem). Filling the remaining gaps in proof sketches
allows to obtain a very detailed cognitive model.

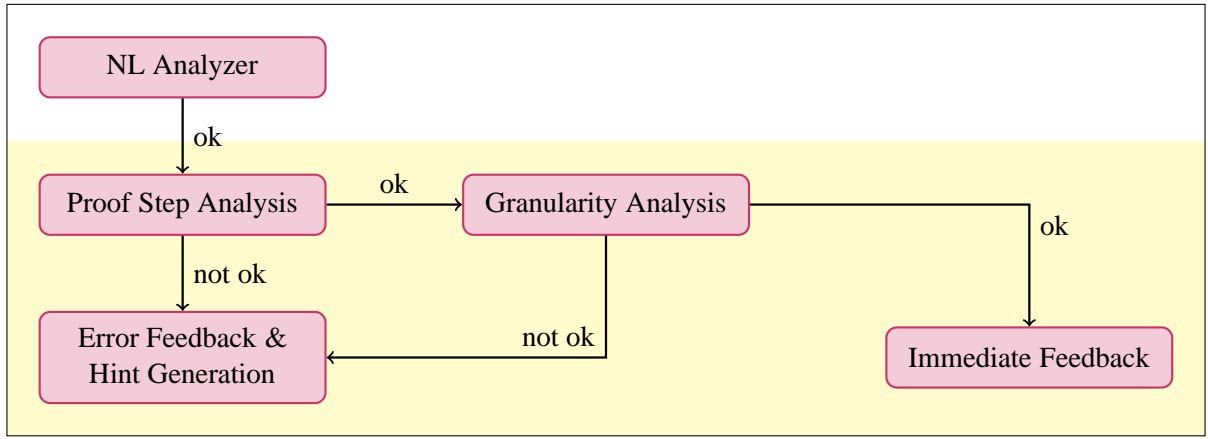
\begin{figure}[tb]
  \centering
  \input{diagram}
  \vspace*{-6mm}
  \caption{Workflow}
  \label{fig:functionalspec}  
\end{figure}

The functional specification of the ITS for proofs is shown in
Figure~\ref{fig:functionalspec}: The student's inputs are analyzed by
a natural language processing module, which provides either a
declarative proof command in the declarative proof script language as
output, or the information that a hint has been requested. The ITS
maintains the focus on the current open goal and interprets the inputs
in that context. Subsequently, the proof analyzer tries to
automatically reconstruct missing proof steps and derive missing
justifications.  If that succeeds, the proof step is further analyzed
with respect to granularity and relevance by the granularity
analyzer. If the reconstruction fails, the proof assistant tries to
find a reconstruction using in addition buggy rules specified by the
tutor. In either case, the result is a feature vector composed of
local proof analysis criteria \emph{soundness}, \emph{granularity},
and \emph{relevance} (see Section~\ref{sec:requirement analysis}). Due
to the characteristics of the solution space, several alternative
proof reconstructions may be possible for the same proof step
sketch. In order to not rule out any of the possible solutions the
student may follow, the ITS must trace all possible reconstructions
simultaneously. All this is described in Section~\ref{sec:step
  analysis}.

In order to generate hints, the ITS shall be able to provide hints
with increasing degree of explicitness. It shall use automated theorem
proving to try to complete the current partial proof (branch) to a
complete proof and use the information used for proof search to
generate the hints. Proof search is conducted by specific tactics,
which the teacher can provide for a specific mathematical domain. The
tactics encode strategic knowledge and the hierarchy of tactics used
to complete a proof is recorded in order to be exploited to generate
hints with increasing degree of explicitness. This is described in
Section~\ref{sec:hint generation}.

\section{Step Analysis}
\label{sec:step analysis}

Human one-on-one tutoring is thought to be effective due
 to its very interactive nature and frequent (step-by-step)
 feedback~\cite{Merrill_Reiser_Ranney_Trafton_1992}. 
 It was found that using step-by-step feedback 
 in an ITS translates to significant 
 learning gains (cf.~\cite{Corbett2001}). 
A recent meta-analysis \cite{vanlehn2011relative} determines that 
 tutoring systems with step-based feedback are almost 
 as effective as human tutoring, and more effective than 
 systems that are answer-based (i.e. they provide feedback at the level 
 of the solution). Interestingly, systems that use even finer levels of 
 feedback (so-called sub-step feedback) are found to be (only) similarly 
 effective to traditional step-based systems.

Implementing the concept of step-wise tutoring requires that, 
whenever the student performs a proof step, the step is evaluated in the 
current context by a step analyzer. In our approach, a  
three-dimensional \emph{feedback vector} is computed with an entry for a 
proof step's soundness,
granularity, and relevance, respectively. Computing the feedback
vector is a two-staged process: First, a \emph{reconstruction
  algorithm} is started that tries to relate the proof step given by
the student to the current proof situation. Afterwards, the derivation
obtained from the reconstruction algorithm is analyzed to compute the
granularity and relevance measure.

When and how the computed feedback is given to the student is
determined by the \emph{feedback policy}. Our default feedback policy
is to give feedback on the correctness of a proof step immediately,
whereas feedback on the granularity and relevance is only given when
the student violates the condition that is demanded by the tutor.

\subsection{Proof Step Reconstruction}
\label{section:proof command process modell}
Didactic considerations require theorem provers to support
\emph{actual mathematical practice}, in addition to providing powerful
automation in a selected mathematical domain. Since the development of
classical automated search based theorem provers and the corresponding
investigations of logical calculi are mainly driven by correctness,
completeness and efficiency issues, these theorem provers operate not
on a ``human-oriented level'', but almost on the ``machine code'' of
some particular logical calculus, such as resolution. Hence, they can
generally not be used as a model tracer. While there exist techniques
to convert (completed) resolution proofs or matrix proofs into natural
deduction proofs, (see for example \cite{andrews80,Miller84}), it
turns out that performing the proof search directly at a more abstract
level is beneficial for the runtime of the reconstruction. Moreover,
it provides the possibility to run in a ``discovery'' mode without
explicitly having to state an isolated proof obligation, which is often
very difficult in a tutorial context due to underspecification or incomplete
information.

\begin{samepage}
\paragraph{Abstract Reasoning: The Assertion Level.}
To come close to the style of proofs as done by humans,
Huang~\cite{huang94reconstructing,Hu-96-a} introduced the \emph{assertion-level},
where individual proof steps are justified by axioms, definitions, or
theorems, or even above at the so-called \emph{proof level}, such as
``by analogy''. The idea of the assertion-level is, for instance, that
given the facts $U\subset V$ and $V \subset W$ we can prove $U \subset
W$ directly using the assertion:
\begin{displaymath}
  \subset_{Trans}:\forall U.\forall V.\forall W. U \subset V \wedge V \subset W \Rightarrow U \subset W
\end{displaymath}
\end{samepage}

An assertion level step usually subsumes several deduction steps in
a standard calculus, say the classical sequent
calculus~\cite{Ge-69-a}. Therefore, traditional theorem provers can
only achieve such conclusions after a number of proof steps. To use an
assertion in the classical sequent calculus, it must be present in the
antecedent of the sequent and be processed by means of decomposition
rules, usually leading to new branches in the derivation tree. Some of
these branches are subsequently closed by means of the axiom rule
which correspond to ``using'' that assertion on known facts or goals.

The technique to obtain such inferences automatically from assertions
follows the introduction and elimination rules of a natural deduction
(ND) calculus~\cite{Ge-69-a} and can be found in \cite{dietrich2011}.

\paragraph{The Reconstruction Algorithm.}
The proof step reconstruction algorithm is based on two main ideas (see \cite{J20} for details):
(i) Represent the possible states the student might be in as
so-called \emph{mental proof state} (MPS). (ii) Given a new proof step
and a MPS, perform a depth-limited BFS at the \emph{assertion level},
trying to derive one/several successor states that are consistent with
the student's utterance, where the consistency is proof command
specific. 
The depth limiter imposes an upper bound on the 
number of assertion level inferences that are 
assumed to be contained implicitly in the student's 
input.\footnote{Note that the correspondence of student steps to
  calculus steps may vary for each calculus.}  
Whether this limit is sufficient depends on 
 (i) the step size of the available proof mechanism 
and (ii) the experience of the student, as we discuss in Section \ref{sec:granularity analysis}.
We have determined such a bound empirically for the corpus of 
students' proof steps from the \woz experiments, as discussed in Section~\ref{sec:evaluation}.
 The bound is needed to guarantee termination of the
reconstruction algorithm, which might otherwise not terminate.\ednote{MS: 
I have replaced this bit of text -- if you disagree please protest! ``The depth limiter specifies how many assertion
steps\footnote{Note that the correspondence of student steps to
  calculus steps may vary for each calculus.} the student is allowed
to perform implicitly. Intuitively we can think of this bound as
reflecting the experience of the student, consequently it should be
based on a student model. Another possibility is to have a fixed bound
and to express the experience of the student only by available facts
in the model. The bound is needed to guarantee termination of the
reconstruction algorithm, which might otherwise not terminate.''\\AS: Ok.}


\begin{figure}[t]
\newcommand{\tfill}{\text{''---''}}
\centering
\begin{tabularx}{\linewidth}{|X|}\hline\vspace*{-7mm}
\begin{minipage}{\linewidth}
{\scriptsize
\begin{mathpar}
  \inferrule*[right={$\textit{Def}=$}]{\inferrule*[right=$\textit{Def}\subset$,rightskip=0.6cm]{\inferrule*[Right=$\textit{Def}^{-1}$]{\inferrule*[Right=$\textit{Def}\circ$]{\inferrule*[Right=$\textit{Def}^{-1}$]{\inferrule*[Right=$\textit{Def}^{-1}$]{\inferrule*[Right=$\textit{Def}\circ$]{\inferrule*[Right=$Ax$]{\;}{\highlight{(x,y)\in S^{-1} \circ R^{-1}}\seq \tfill}}{\highlight{(z,y)\in R^{-1} \wedge (x,z)\in S^{-1}} \seq \tfill}}{(y,z)\in R \wedge (x,z)\in S^{-1} \seq \tfill}}{\highlight{(y,z)\in R \wedge (z,x) \in S} \seq \tfill}}{(y,x) \in (R \circ S) \seq \tfill}}{\highlight{(x,y)\in (R \circ S)^{-1}} \seq (x,y) \in S^{-1}\circ R^{-1}}}{(R\circ S)^{-1} \subset S^{-1} \circ R^{-1}} \\ 
\inferrule*[Right=$\textit{Def}\subset$]{\inferrule*[Right=$\textit{Def}\circ$]{\inferrule*[Right=$\textit{Def}^{-1}$]{\inferrule*[Right=$\textit{Def}^{-1}$]{\inferrule*[Right=$\textit{Def}\circ$]{\inferrule*[Right=$\textit{Def}^{-1}$]{\inferrule*[Right=$Ax$]{\;}{\highlight{(x,y)\in (R\circ S)^{-1}}\seq \tfill}}{\highlight{(y,x)\in (R\circ S)} \seq \tfill}}{\highlight{(z,x)\in S \wedge (y,z)\in R} \seq \tfill}}{(x,z) \in S^{-1} \wedge (y,z)\in R \seq \tfill}}{\highlight{(x,z)\in S^{-1} \wedge (z,y)\in R^{-1}} \seq \tfill}}{\highlight{(x,y)\in S^{-1}\circ R^{-1}} \seq (x,y) \in (R\circ S)^{-1}}}{S^{-1} \circ R^{-1} \subset(R\circ S)^{-1}}
}{\vdash (R \circ S)^{-1} = S^{-1}\circ R^{-1}}
\end{mathpar}
}\vspace*{-7mm}
\end{minipage}\\\hline
\end{tabularx}
\caption{Annotated \OMEGA assertion level proof}
\label{fig:reconstruction}
\end{figure} 

Figure \ref{fig:reconstruction} shows an example reconstruction of a
complete dialog taken from the corpus. In the figure, the shaded
formulas correspond to the steps entered by the student. The white
formulas correspond to assertions the student has left out and which
were filled in by the reconstruction module.

A MPS is represented as a set of sequents that are the subproblems to be
solved, together with a global substitution which instantiates
meta-variables. One of these sequents is always marked and represents the
sequent the student is working on. 
Always keeping track of the student's subgoals facilitates task sensitive feedback.

Initially, the MPS is unique and consists of the exercise given to the
student as single sequent, together with the empty substitution. During
the search, an invariant is that a MPS always represents a valid proof
state. By expanding a given proof state only by valid actions, it is
guaranteed that only reachable and consistent proof states are
generated.  Let us stress again that due to ambiguity and
underspecification several consistent successor states are possible
(as in the case of statement {\bf S8b} shown in
Figure~\ref{corpus:st25p2} where the next subgoal the student will
work on is underspecified). There can also be several reconstructions
for a given proof step. Therefore, the verification algorithm works on
a list of MPS rather than on a single one.

While the reconstruction algorithm might look similar to the
processing model of proof commands in a pure verification setting,
there are the following subtle differences:

\begin{itemize}
  \item In a pure verification setting, it is sufficient to find some
    verification for a proof command. The verification itself is
    usually not of interest and needs not to be further processed. In
    contrast, in a tutorial setting we need to consider several, if
    not all, possible verifications of the given proof command and
    need to relate them to the student's knowledge to avoid the
    student to rely on the power of the underlying theorem prover to
    solve the exercise.
  \item In a pure verification setting, we can assume the user to be
    an expert in the problem domain as well as in the field of formal
    reasoning. This has several implications 
    on the processing model:
    (i) inputs can be expected to be correct and just need to be
    checked, (ii) proof commands can lazily be verified until a (sub)proof is
    completed, (iii) justification hints are given that indicate how
    to verify a given proof command, (iv) feedback is limited to
    ``checkable'' or ``not checkable''.
    In contrast, in a tutorial setting, we must assume the user to be neither a domain
    expert nor an expert in formal reasoning. The underlying
    mechanisms need to be hidden from the user, direct and
    comprehensive feedback has to be provided at each step. Therefore,
    it is for example a requirement to anticipate why an assumption
    is made, in contrast to a lazy checking once the conclusion has
    been obtained.
  \item In a pure verification setting, we can assume the user to
    indicate when the proof of a subgoal is finished (as usually done
    by so-called \emph{proof step markers} in the proof
    language). However, in the tutorial setting this information is
    implicit. Similarly, we must be able to perform backward steps
    where some of the new proof obligations have not yet been shown.
\end{itemize}

In order to illustrate how the verification algorithm works, we will
step through the verification of utterance {\bf S8} from
Figure~\ref{corpus:st25p2}, beginning with the initial MPS and
finishing with the MPS extended by the proof step. The initial MPS is
$\{ \langle \underline{\vdash (R\circ S)^{-1}=S^{-1}\circ R^{-1}} ;
\emptyset \rangle \}$ and the proof step to be verified is
$\mathbf{let}\; (x, y)\in (R \circ S)^{-1}$.

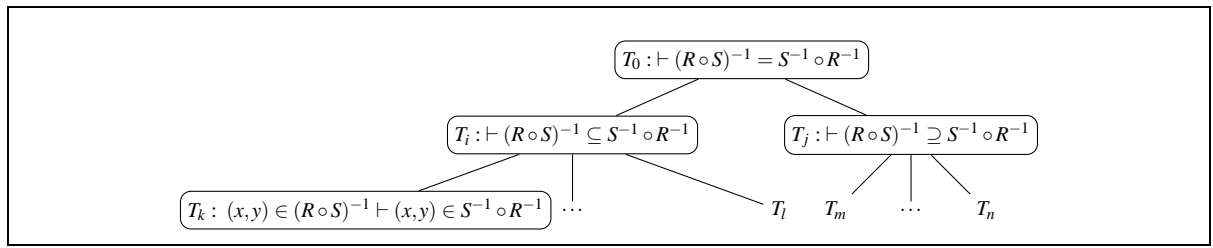
\begin{figure}[b]
\begin{tabularx}{\linewidth}{|X|}\hline\vspace*{1mm}
\begin{minipage}{\linewidth}
\scriptsize
\tikzstyle{pnode} = [rectangle, rounded corners, draw]
\centering
\begin{tikzpicture}[sibling distance=45mm, level distance = 1cm]
  \node [pnode] (z){$T_0:\;\vdash (R\circ S)^{-1}=S^{-1}\circ R^{-1}$}
   child { 
     node [pnode] {$T_i:\;\vdash (R\circ S)^{-1}\subseteq S^{-1}\circ R^{-1}$}
     [sibling distance=2.75cm]
     child { node [pnode] {$T_k:\;(x,y)\in (R\circ S)^{-1}\vdash(x,y)\in S^{-1}\circ R^{-1}$} }     
     child { node {$\ldots$}}
     child { node {$T_l$}}     
    }
   child {
     node [pnode] {$T_j:\;\vdash (R\circ S)^{-1} \supseteq S^{-1}\circ R^{-1}$}
     [sibling distance=1cm]
     child { node {$T_m$}}     
     child { node {$\ldots$}}
     child { node {$T_n$}}     
     }
    ;    
\end{tikzpicture}
\end{minipage}\vspace*{2mm}\\\hline
\end{tabularx}
\caption{The expanded proof state after step (i) of verification (abbreviated).}
\label{example:step1}
\end{figure}

Having expanded the current proof state (step (i), shown in
Figure~\ref{example:step1}), we apply a \textbf{let}-proof step specific filter 
to find the set of newly-created sequents which are consistent with the
given proof step.
Of the sequents in the tree, only the node containing the sequent $T_k$
passes, since the formula in the proof step appears on the left-hand
side of the sequent. Now that we have found the consistent successor
sequents, we must complete these sequents to MPSs. Because the
decomposition of the sequent $T_0$ introduced a subgoal split, the sequent
$T_j$ must be proved in addition to $T_k$. The resulting MPS is
therefore $\{ \langle \underline{T_k},T_j ; \emptyset \rangle \}$,
that is, $T_k$ is now the current sequent, and $T_j$ is still to be
proved. Finally, we prune the nodes which were rejected
by the filter.

\paragraph{Relevance Checking.}

Using the proof step reconstruction mechanism for each proof step allows 
our approach to perform a form of relevance checking when a 
hypothesis is introduced. A hypothesis introduced by the student 
is matched against a proof search in \OMEGA, and considered relevant 
only if it can be unified with a step that is part of one of the 
partial solutions that are discovered via strategic proof search.

\medskip
In practice, it turns out that it is very important for a tutoring
system 
to enable a broad range of people to 
create content for the
system in form of exercises and domain expertise. One of the main
advantages of our approach is to use existing mature representation
and search technology that has been developed over the last decades in
the context of ITP/ATP. New domains can easily be added by users either
by relying on already existing specifications of formalized
mathematics, or by writing new specifications from scratch. That is,
the only information the author has to provide is a problem
description and the knowledge needed to solve the problem. As
inferences are automatically synthesized from theorems and
definitions, it is sufficient to provide this knowledge in a
declarative form. For simple domains, this is already sufficient and
there is a high chance that modifications of existing proofs or even
new proofs are recognized by the tutor. For more complex domains in
which the reasoning is more complicated, the author also has to
provide strategic information on how to solve a problem.

\subsection{Granularity Analysis} \label{sec:proof step analysis}
 \label{sec:granularity analysis}

In addition to verifying the correctness of proof steps generated by the 
student, and to detect steps that are logically incorrect, we use 
proof reconstructions to judge about another qualitative aspect of 
proof steps: granularity. By assessing the step size (in the context of 
the ongoing proof and a student model), 
a tutoring system for proofs can 
react if the student's solution lacks necessary detail, or, to the contrary, 
the student is progressing at smaller steps than expected, and adapt 
feedback and hints accordingly. Having a metric for step size also allows 
the system to generate and present hierarchical proofs (or steps to be used as hints) 
at specific levels of granularity.



We have devised a framework to analyze the step size of 
proof steps \cite{marvinAKA2010}, 
where a proof step can refer to either the single application 
of an inference rule, or consist of several (tacit) intermediate 
inference applications provided by the reconstruction algorithm.
Granularity judgments for such a (single or aggregate) step are
considered as a classification task. Proof steps are characterized
according to a catalog of criteria (cf.~\cite{marvinAKA2010}) that are
thought to be indicative of granularity, and classified as
appropriate, too small, or to big according to a classifier.
 
\begin{figure}
\centering
\tikzstyle{test}=[shape aspect=3,diamond,draw,fill=black!25,font=\slshape]
\tikzstyle{decision}=[rectangle,draw]
\begin{tabularx}{\linewidth}{|X|}\hline\vspace*{1mm}
\begin{minipage}{\linewidth}\centering
\begin{tikzpicture}[node distance=12mm]
  \node[test] (total) {\makebox(13,9){total}};
  \node[test,below right=of total] (mcu) {\makebox(13,9){m.c.u}};
  \node[decision,below left=of total] (app1) {app.};
  \node[test,below=5mm of mcu] (relations) {\makebox(14,10){relations}};
  \node[test,left=of relations] (hypintro) {\makebox(14,10){hypintro}};
  \node[decision,right=of relations] (bg1) {too big};
  \node[decision,below left=10mm of hypintro] (sm1) {too small};
  \node[decision,right=20mm of sm1] (app2) {app.};
  \node[decision,below right=10mm of relations] (bg2) {too big};
  \node[decision,left=20mm of bg2] (app3) {app.};
  \draw (total) -- node[right=5pt,pos=.3] {$>1$} (mcu); 
  \draw (total) -- node[left=5pt,pos=.3] {$\leq 1$} (app1); 
  \draw (mcu) -- node[left=9pt,pos=.3] {$>0 \wedge \leq 2$} (hypintro); 
  \draw (mcu) -- node[right] {$3$} (relations); 
  \draw (mcu) -- node[right=9pt,pos=.3] {$>3$} (bg1); 
  \draw (hypintro) -- node[left=5pt,pos=.3] {$=0$} (sm1); 
  \draw (hypintro) -- node[right=5pt,pos=.3] {$>0$} (app2); 
  \draw (relations) -- node[left=5pt,pos=.3] {$\leq 2$} (app3); 
  \draw (relations) -- node[right=8pt,pos=.3] {$3$} (bg2); 
\end{tikzpicture}
\end{minipage}\vspace*{2mm}
\\\hline
\end{tabularx}
\caption{Example granularity classifier}
\label{granularityclassifier}
\end{figure}
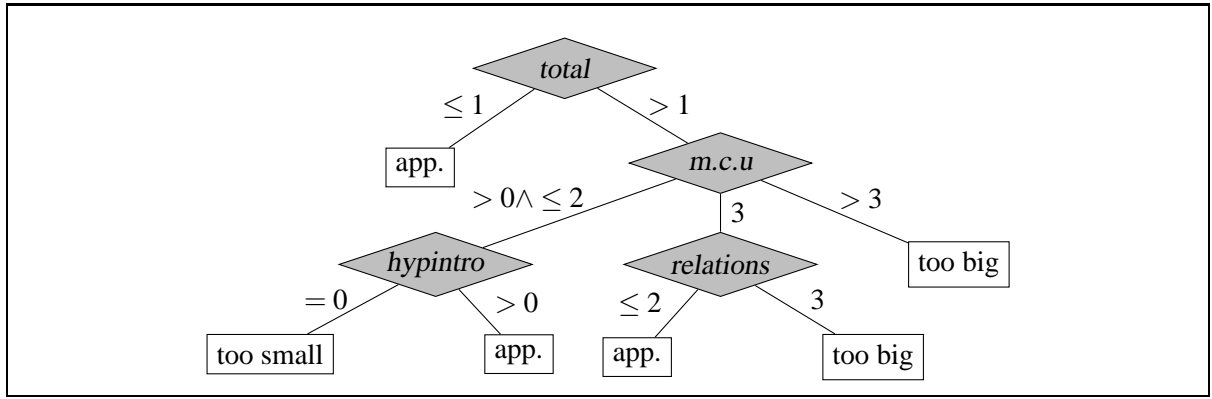

Granularity criteria, which are the basis for the classification task,
take into account the current proof context, the content of the
student model, and the verbal explanations given by the student. We
currently use an overlay student model which for each assertion-level
inference rule maintains an assumption whether it is mastered by the
student or not, based on the student's actions.\footnote{There are
  more sophisticated techniques (e.g.~Bayesian networks) for
  estimating the student's knowledge that can be used instead.
  However, student modelling as such was not the focus of this
  research.}  Analyzing a proof step with respect to the catalog of
criteria yields a vector that is encoded numerically.  For example,
the criterion ``unmastered concepts'' is assigned the count of
mathematical concepts employed as inferences in the (simple or
aggregate) step which are supposed to be not yet mastered by the
student.  Classifiers can be represented in the form of decision
trees, where decision nodes represent granularity criteria, and leaves
record the granularity verdict. In our evaluation discussed in 
Section~\ref{sec:evaluation}, we have also considered other forms of 
classifiers, such as rule based classifiers and classifiers learned by
the support vector machine approach. An example for a simple decision
tree classifier for granularity is presented in
Figure~\ref{granularityclassifier}. According to this particular
classifier, for example, a proof step that consists of two inference
applications at the assertion level (total=2), which represent two
different concepts both of which have previously been mastered
according to the current state of the student model (m.c.u.=2) and
none of which introduces a new hypothesis into the proof (hypintro=0)
is classified as ``too small''. Note that this particular decision
tree only uses a small number of criteria.  Granularity classifiers
can be written by hand, but an interesting question is what kind of
judgments human experts actually make when assessing proof steps.  In
order to assess what granularity criteria are relevant for human
tutors, and whether corresponding classifiers can be learned from
samples of proof steps annotated with granularity judgments, we
conducted an experiment presented in Section \ref{sec:evaluation}.

\section{Next Step Generation for Hinting}
\label{sec:hint generation}
At any time of a tutorial session, a student might get stuck and
request help on what step to perform next. Help can also be given
without an explicit request, for example after repeated student
errors, or a long period of silence. Thus, one important design
decision for a tutoring system is \emph{when} to give a hint, i.e., to
provide a \emph{hinting policy}. 
\cbstart[4pt]
For our tutoring system, we use the
simplest possible hinting policy, namely to give a hint only if it is
explicitly requested by the student. As discussed in
\cite{Aleven2000}, this might not be optimal, as students might abuse
this functionality or refuse to ask the system for a hint;
nevertheless it is the strategy that is used in most tutoring
systems. 

It has been shown that human tutors use hint sequences 
that start with abstract hints and refine the hint on demand. 
The principle of progressively providing more concrete hints  
if required has been applied to a number of tutoring systems, including 
 the Carnegie Proof Lab \cite{DBLP:journals/igpl/Sieg07}. 
 The goal is to leave the student to
 perform the actual concrete steps that the hint has requested, 
 leading to better knowledge construction. 
  If the student is still stuck, subsequent hints
  should refer to smaller subtasks of the proof, becoming increasingly close to
 the fully-specified assertion level step.

To find a relevant, context-sensitive hint, we follow the typical
approach of MTTs and invoke the domain reasoner to find a solution for
the current proof state. This solution is then analyzed to extract a
hint. 
\cbend
In our approach, the provision of increasingly concrete hints is supported 
by a problem solver that generates
a \emph{hierarchical solution} (see \cite{ABDMW-05-a} for details) based
on proof strategies, where each (sub)invocation of a strategy
introduces a new hierarchy in the computed solution. Intuitively, a
high level in the hierarchy sketches how the overall
problem was structured into subproblems. At the lowest
level, a concrete proof with concrete assertion steps is given.
Consequently, we can synthesize both \emph{strategic hints} as well as
information about the \emph{concrete next step} to be performed. Of
course, the quality of the hints directly depends on the quality of
the proof strategies, i.e., how the knowledge is encoded by the author
of the exercise.


We first describe in Section \ref{sec:proof strategies} how an author
can encode proof strategies and how these strategies are used to
generate a hierarchical proof object, and in Section \ref{sec:hinting}
how the computed solution is used to synthesize a hint.

\subsection{Authoring of Proof Strategies}\label{sec:proof strategies}
A \emph{proof strategy} represents some mathematical technique that
happens to be typical for a given problem. For example, there are
strategies which perform proof by induction, proof by contradiction,
solve equations, or unfold definitions. To achieve a (strategic) goal,
a strategy performs a heuristically guided search using a dynamic set
of assertions, as well as other strategies.

In contrast to other approaches that require to encode the knowledge
in the underlying programming language of the system, we encode proof
strategies in a separate strategy language (see
\cite{crystal-JAL,DBLP:conf/itp/AutexierD10} for an
overview). 

A simple proof strategy that is proposed in \cite{Solow05} is the
``Forward-Backward Method'', which combines the two well-known problem
solving strategies: \emph{forward chaining} and \emph{backward
  chaining}: The method starts with backward chaining by matching the
current proof goal with the conclusions of theorems and definitions and
adding their premises as new goals to be proved. The backward chaining
phase continues until all conclusions have been solved or until no
further definition or theorem can be applied. Subsequently, a forward
chaining phase is started. Forward chaining starts from available
assumptions and given facts and continuously applies definitions and
theorems forwards by instantiating all their premises and adding their
conclusions to the current proof state.

We have formalized this method as a strategy ``Close-by-Definition''.
Each phase is realized by a sub-strategy: ``Work-Backward'' 
applies definitions in backward direction, as indicated by the keyword
{\bf backward}, that is, expands definitions that occur in the
goal. ``Work-Forward'' applies all definitions in forward direction,
as indicated by the keyword {\bf forward} (see Figure~\ref{fig:strategies}), 
that is, expands concepts that occur on the
left-hand side of the sequent. It is the responsibility of the author
of the strategy to guarantee termination.

As additional logical steps are commonly needed to close a proof task,
we provide a third strategy ``Close-by-Logic'', which performs
case-splits and applies the axiom rule to close sequents. Finally,
these strategies are assembled to the overall strategy
``Close-by-Definition'', which calls the strategies in a specific
order. The strategy keyword {\bf try} ensures that a strategy
application can also be skipped, e.g., when the student has already
expanded the goal completely.

\begin{figure}[t]
\fbox{
\begin{minipage}{.98\linewidth}
\begin{minipage}{0.45\linewidth}
\begin{isabellebody}
\isacommand{strategy} work-backward\isanewline
\ \  \isacommand{repeat} \isanewline
\ \  \isacommand{use} \isacommand{select} * \isacommand{from} definitions \isacommand{as} \isacommand{backward}\isanewline
\end{isabellebody}
\begin{isabellebody}
\isacommand{strategy} close-by-logic\isanewline
\ \  \isacommand{repeat} \isacommand{first} deepaxiom, or-l
\end{isabellebody}
\end{minipage}
\hspace{0.1cm}\vline\hfill
\begin{minipage}{0.5\linewidth}
\begin{isabellebody}
\isacommand{strategy} work-forward\isanewline
\ \  \isacommand{repeat} \isanewline
\ \  \isacommand{use} \isacommand{select} * \isacommand{from} definitions \isacommand{as} \isacommand{forward}\isanewline
\end{isabellebody}
\begin{isabellebody}
\isacommand{strategy} close-by-definition\isanewline
\ \ \isacommand{try} work-backward \isacommand{then} \isacommand{try} work-forward \isacommand{then} close-by-logic
\end{isabellebody}
\end{minipage}
\end{minipage}}
\caption{Formalization of a simple proof strategy}
\label{fig:strategies}
\end{figure}

\subsection{Hinting}\label{sec:hinting}
Once a solution of the current proof state has been computed in the
form of a hierarchical proof, the proof hierarchies can be used to
synthesize hints of increasing specificity. This is done as follows:
(i) selecting a certain level of hierarchy in the reconstruction, (ii)
selecting a successor state at the selected hierarchy (iii) extracting
information from the selected state and converting it to a concrete
hint. A given hint can be refined by either switching to a more
detailed level in the hierarchy, or by increasing the information which was
extracted from the selected successor state. 

How should the next proof step be communicated to the user? 
A general issue in ITS design is how much
 scaffolding is to be provided to the learner, which is known as the 
``assistance dilemma'' -- both too much and 
too little assistance hamper learning~\cite{KoedingerAndAleven2007}.
So-called \emph{Socratic} teaching strategies (cf.~\cite{Rose01acomparative}) have been
demonstrated to be superior to \emph{didactic} teaching strategies, 
i.e. hinting based on direct instruction,  
especially regarding their long-term effects \cite{Rose01acomparative,Chi94elicitingself-explanations,Ashley02teachingcase-based}.
Socratic teaching is motivated by the 
idea that learners a priori posses the necessary prerequisites to 
acquire new knowledge from existing knowledge (cf.~\cite{Woolf2008}). The role of the tutor thus 
is to moderate this process by asking knowledge-eliciting questions. 
Such a teaching strategy for 
the domain of mathematical proofs has been developed and automated 
 by Tsovaltzi~\cite{dimitraDiss} and provides the background for our work.\ednote{Question to all: may we claim this? Is there another nice word that explains that our work is inspired by this, but not exactly congruent?}
By applying the Socratic teaching strategy, the next proof
step is not given directly to the student. Instead, a
question is formulated that encourages the student to think for
himself and construct his own solution to the task at hand. 

Similar to \ANDES (see \cite{Gertner98proceduralhelp}), we use
templates to generate hints in natural language. Variables in the
templates are filled with the concrete objects that are available in
the actual proof situation. We have designed a general ordered set of
templates that can be used for arbitrary assertion level steps;
moreover, we attach an ordered set of templates to each strategy.
The hints can be classified as follows:

\begin{enumerate}[(i)]
  \item A strategic hint that describes 
 what to apply from a strategic viewpoint, 
for example: ``What assertion can be applied backward to the
    goal?''
  \item A hint on an inference to be applied by pointing to involved
    variables, for example: ``Can you say anything about the sets $A$ and $B$?''
  \item A hint on an inference to be applied without stating the 
    premises and conclusion that are involved, for example: ``How can
    you show that two sets are equal?''
  \item A hint that points to the premises of an inference that is
    applied backwards without naming the inference, for
    example: ``If you want to show that $A \cap B = B\cap A$, what
    should be true about these sets?''
  \item A hint that points to a subgoal but does not say how the
    subgoal is achieved, for example: ``How can you show that $A \cap
    B \subset B \cap A$?''
  \item A hint that points to the conclusion of an inference that is
    applied forwards without naming the inference, for
    example: ``What can you conclude if you know that $x \in A$ and $x \in B$?''
  \item A hint that describes the complete application of an
    inference, that is, the name of the inference, together with the instantiated premises and conclusion.
  \item A hint that points to an inference application together with
    restating the assertion to be applied.
\end{enumerate}

\begin{figure}[t]
  \centering
  \begin{tabularx}{\linewidth}{|X|}\hline
  \begin{minipage}{\linewidth}
  \tikzstyle{strategyname}=[font=\scriptsize]
  \begin{tikzpicture}[node distance=10mm]
    \node[] (t0) {};
    \node[circle,draw,right=of t0] (t1) {$T_1$};
    \node[circle,draw,below right=of t1] (t2) {$T_2$};
    \node[circle,draw,right=of t2] (t3) {$T_3$};
    \node[circle,draw,above right=of t3] (t4) {$T_4$};
    \node[below right=6mm of t4] (t8) {\dots};
    \node[circle,draw,right=30mm of t4] (t5) {$T_5$};
    \node[circle,draw,right=30mm of t5] (t6) {};
    \node[below right=6mm of t5] (t7) {\dots};
    \draw[dashed,arrows=->] (t0) -- (t1); 
    \draw[solid,arrows=->] (t1) -- node[right,pos=.4,strategyname] {Def $\subset$} (t2); 
    \draw[dashed,arrows=->] (t2) -- (t3); 
    \draw[solid,arrows=->] (t1) -- node[above,midway,strategyname] {Work-Backward} (t4); 
    \draw[solid,arrows=->] (t3) -- (t4); 
    \draw[solid,arrows=->] (t4) -- node[above,midway,strategyname] {Work-Forward} (t5); 
    \draw[solid,arrows=->] (t5) -- node[above,pos=.35,strategyname] {Close-by-Logic} (t6); 
    \draw[solid,arrows=->] (t5) -- (t7); 
    \draw[solid,arrows=->] (t4) -- (t8); 
    \draw[solid,arrows=->] (t1) to [bend left=20] node[above,midway,strategyname] {Close-by-Definition} (t6); 
  \end{tikzpicture}
  \end{minipage}\vspace*{1mm}\\\hline
\end{tabularx}
\caption{Hierarchical proof plan completing the proof of task $T_1$}
  \label{fig:proof}
\end{figure}
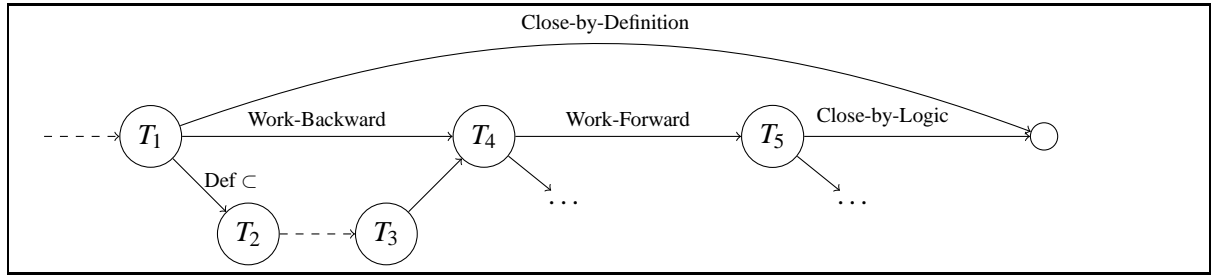

Let us illustrate our approach by means of an example. Consider the
exercise $(R \circ S)^{-1} = S^{-1}\circ R^{-1}$. Suppose that the
student starts the proof by 
equality, yielding two subtasks
\begin{align}
  T_1 :  &\vdash (R \circ S)^{-1} \subset S^{-1}\circ R^{-1} \\
  T_1': & \vdash S^{-1}\circ R^{-1} \subset (R \circ S)^{-1} 
\end{align}
and requests a hint for the task $T_1$. A possible completion of the
proof, encoded in the strategy ``Close-by-Definition'', consists of
expanding all definitions and then using logical reasoning to complete
the proof. The resulting hierarchical proof object is
shown schematically in Figure~\ref{fig:proof}. The task $T_1$ has
three outgoing edges, the topmost two corresponding to a strategy
application and the lower-most 
one 
corresponding to an assertion
application, respectively. Internally, the edges are ordered with
respect to their granularity,  
according to the hierarchy of nested strategy applications 
 that generated them
. In the example the most abstract
outgoing edge of $T_1$ is the edge labelled with
``Close-by-Definition'', followed by the edge labelled with
``Work-Backward'', both representing strategy applications. The edge
with the most fine-grained granularity is the edge labelled with ``Def
$\subset$'' and represents an inference application. 

By selecting the edges ``Work-Backward'', ``Work-Forward'', and
``Close-by-Logic'', we obtain a flat graph connecting the nodes $T_1$,
$T_4$, and $T_5$. A more detailed proof-view can be obtained by
selecting the edge ``Def $\subset$'' instead of ``Work-Backward''. In
this case the previous single step leading from $T_1$ to $T_4$ is
replaced by the subgraph traversing 
$T_2$ and $T_3$. 

As mentioned above, each selection can be used to generate several
hints. Suppose for example that we select the lowest level of
granularity, and the first proof state to extract a hint. This already
allows the generation of three hints, such as
  \begin{itemize}
  \item{``Try to apply Def $\subset$''}
  \item{
``Try to apply Def $\subset$ on  $(R \circ S)^{-1} \subset S^{-1}\circ R^{-1}$'' 
}
  \item{
``By the application of Def $\subset$ we obtain the new goal $(x,y) \in  (R \circ S)^{-1} \Rightarrow  (x,y) \in S^{-1}\circ R^{-1}$" 
}
  \end{itemize}
Selecting a more abstract level would result in hints like ``Try to
work backward from the goal'', or ``Try to apply definitions on the
goal and assumptions''. 
The ability to provide hints that address several 
dimensions at which scaffolding can be provided in proof tutoring 
is beneficial for providing targeted, adaptive feedback.

\section{Evaluation}
\label{sec:evaluation}

The presented techniques have been successfully evaluated 
using experiments and experimental data from the 
\woz experiments described in Section~\ref{sec:woz study}. 
In particular, we examined in how far the proof reconstruction 
mechanism successfully models the proofs from the students 
(who were unconstrained in their solution attempts). We performed 
additional experiments to study granularity judgments by human 
tutors, and in how far these judgments can be learned via machine 
learning techniques and represented as classifiers within 
the tutoring system.

\paragraph{Proof Reconstruction \& Assessment.}

In the \woz experiments 
introduced in Section~\ref{sec:woz study}, tutors were asked to indicate explicitly 
whether steps are correct or incorrect. We investigated in how far 
the steps judged as correct by the tutors are reconstructed by our 
approach -- to provide appropriate feedback on correctness, but also 
to serve as the basis for further analysis, e.g.~granularity analysis. 
Since the algorithm uses breadth-first-search, an important question 
is what search depth is necessary to verify the proof steps from the 
students.

In our analysis, we used 144 proof steps from the \woz experiment 
which deal with an exercise 
about binary relations, namely to show 
that $(R \circ S)^{-1} = R^{-1} \circ S^{-1}$ (where the operators $\circ$
  and $^{-1}$ denote relation composition and inverse). 
Of these steps, 116 were judged 
as correct and 28 as incorrect. Table \ref{rec-eva-results} shows the 
proportion of steps that were correctly accepted and rejected, and 
the proportion of steps that were not verified or wrongly accepted, using 
a depth-limit of four steps for BFS. 
Apart from three steps, all proof steps are correctly verified or 
rejected (which corresponds to an overall accuracy of 98\%). 
Since the relatively small depth limit of
 four steps is sufficient for accurate verification in our sample, 
we conclude that our approach to proof reconstruction is feasible 
(within the given domain of proofs).

\begin{table}

\begin{center}
\begin{tabular}{lll}
\toprule
& Verified & Rejected    \\
\midrule
Step correct   & {\bf 113 (97\%)} & 3 (3\%) \\
Step incorrect & 0                & {\bf 28 (100\%)}  \\
\bottomrule
\end{tabular}
\end{center}

\caption{Number of steps that were verified or rejected via 
proof reconstruction. Bold numbers indicate accurate classification, 
relative to the true class of the steps (correct/incorrect).}
\label{rec-eva-results}
\end{table}

\paragraph{Granularity.}

We have used proof reconstructions at the assertion level 
as the basis for granularity analysis 
(as outlined in Section \ref{sec:granularity analysis}) 
within a framework for judging granularity 
(presented in \cite{marvinAKA2010}). 
To investigate the prospect of learning granularity classifiers 
from the granularity judgments by expert tutors, 
we asked four expert tutors to contribute to a corpus 
of granularity judgments. This corpus was used to analyze 
agreement between tutors, 
to synthesize granularity classifiers via machine learning 
techniques, and to determine what granularity criteria are most useful 
for judging granularity (as presented in more detail in \cite{marvinAKA2010}). 
A first exploratory analysis revealed that 
most of the proof steps that were presented to the tutors 
(which were constructed from one or a few inference steps at the 
assertion level) were considered to be of appropriate granularity. 
This is in particular the case for proof steps that correspond to 
one single inference at the assertion level, which were considered 
of appropriate size in 93\%, 69\%, 83\% and 96\% of the cases 
by the four tutors. When the bias towards the ``appropriate'' 
class is accounted for, agreement 
between tutors is moderate.  
 The individual sub-corpora of granularity judgments by the four tutors also differed 
 in how far they were found to be amenable to the automated learning 
 of classifiers. We used several algorithms (decision tree learning, 
 decision rule learning and support vector machines) offered by the 
 data mining tool Weka\footnote{http://www.cs.waikato.ac.nz/ml/weka/}, 
 with different degrees of success for the four sub-corpora (cf.~\cite{marvinAKA2010}).
It should be noted that 
the experiment investigated the naturalistic judgments of tutors 
without enforcing ``consistency'' 
in the judgments, therefore some disagreement was to be expected.
Overall, the experiments indicated that the proof steps at the assertion 
level are a good basis for granularity analysis, since they are 
close to what human judges (in the setting of our experiments) 
consider as appropriate step size. Furthermore, counting the number 
of assertion level steps in the reconstruction of a student's proof 
step was determined as the most indicative criterion when only judgments 
are considered where three out of the four judges agree.
 The experiment illustrates 
the prospects of learning classifiers from expert tutors, but it also points 
at the differences between experts in judging granularity.

\paragraph{Discussion.}

The evaluation illustrates the use of assertion level reasoning 
for assessing proof steps as they occur in relatively unconstrained 
tutoring dialogs that we collected in the \woz experiments.
Furthermore, we have shown that these reconstructions are a useful 
basis for analyzing further aspects of proof steps, such as granularity.



\section{Outlook}
\label{sec:future} 

The previous sections describe how the proof assistant \OMEGA was
utilized as a domain reasoner providing feedback on proof steps and
generating hints. We believe that using a proof assistant offers a lot
more of not yet exploited potential to improve the quality of 
  ITSs.
In this section we present some of these ideas that are the basis for
future work.

\paragraph{Further Qualitative Proof Step Assessment Criteria.} 
Reconstruction of a proof step is usually a task that is \emph{local} to
an individual student session and can be seen as the process of
generating an individual solution graph for the given subject. To make
the tutoring system more efficient, it is possible to cache solution
graphs from previous tutoring sessions and to combine them to an
overall solution graph. This has two major advantages: (i) The
instructor gets a compact overview over typical approaches taken by
all of his students, (ii) the instructor can review and refine hints
that were given by the system, (iii) each student that comes up with a
new solution becomes an author, (iv) whether a proof is sensible or
whether a (irrelevant) proof step is nevertheless sensible typically
are properties that need the comparison with other solutions. The hope
is that eventually a fixed point will be reached, containing a fully 
specified trace over all possibilities.

Another criterion to analyze a proof step for is whether it is
consistent with an overall strategy. For instance, if the teaching
goal also consists in teaching a specific proof strategy, then the proof
assistant must interpret a student's proof step not only with respect
to a current open goal, but also with respect to an upper strategy.
In case a strategy consists of different sequential sub-strategies,
such as, for instance, the ``Forward-Backward Method'' from
Section~\ref{sec:proof strategies}, this also requires to be able to
find out when one strategy is finished and the next one starts. How to
achieve this tracking of strategy execution is an open problem, but a
solution could be to hook the user input into the tactic execution
mechanism.

\paragraph{Integration of Model Generators.}
A variety of tools has been developed for finding finite models of
first order logic (FOL) formulas, such as
\PARADOX~\cite{Claessen03newtechniques} and 
\MACE~\cite{DBLP:journals/corr/cs-SC-0310055}, to name a few. Given for
example the theory of groups and the assertion that all groups are
commutative, they are able to produce a counter-model of the
assertion, i.e., a non-commutative group. This is done by providing an
interpretation for the involved function symbols which makes the
assertion false, which can be understood as a group table. Similarly,
counter-models can be generated for other theories, such as the theory
of binary relations. Note that in contrast to the verification of a
proof step, the counter-model provides the information that the
given step is wrong. 
Proof step reconstruction could be made more efficient by
systematically checking for counter-models during the reconstruction
process. Moreover, in case no reconstruction was possible,
model-finders could be employed to generate a counter-model for the
proof step and the error-feedback routine could employ the
counter-model to provide hints or explain why a proof step is
invalid. The challenge here consists of adequately verbalizing the
found counter-model.

\paragraph{Reviewing of Solution (service S\ref{service:solution
    review}, p.~\pageref{service:solution review}).} Mathematical
proofs in textbook-style mainly contain forward proof steps. However,
to find a proof often a backward-style is used.  Checking if a
specific proof step entered by the student is in forward-style or in
backward-style is easy. Depending on the didactic strategy and
possibly depending on the skills of a student, the ITS has the choice of enforcing forward-style proof steps
immediately, or to let a student having difficulties with proving in
the first place write a proof in any style. In the second case, the
tutor can then review the solution with the student, for instance, by
indicating backward proof steps and subsequently asking to transform them 
into a forward-style proof. Or simply by showing the student his own
proof in forward-style, which is easily possible in \OMEGA already
now. 

\paragraph{Assessment of Student Knowledge (service S\ref{service:knowledge
    assessment}, p.~\pageref{service:knowledge assessment}).}
\cbstart[4pt]
The student's knowledge is incorporated in the granularity classifier
by exploiting the information which concepts are mastered by the
student. During the proof development that information 
 is 
updated, for instance, by adding initially unmastered concepts to the
mastered concepts once they have been used correctly a number of times
in a proof step. 
Since our approach uses a full-fledged proof assistant system 
for the analysis of the student's input, a precise and detailed 
assessment of the student's actions is provided, which is 
considered beneficial for student modeling. In this context, the benefit of 
using state-of-the-art techniques in student modeling 
for diagnosing student knowledge 
(using, for example, statistical inference) could be explored. 
Such a modeling of student knowledge can enable the system to adjust 
instructions and exercises to the particular strengths and difficulties 
of individual students.

Furthermore, an interesting feature that an ITS as outlined in this paper 
could offer is a walk-through of the student's proof solution,  
 where the proof steps that were 
rejected by the system are presented along with the reasons for rejection, 
as well as the accepted proof steps and their relation to the final proof. 

\cbend

\paragraph{Automatic hint generation for logic proof tutoring using
  historical data.}  So far the hint generation uses the recorded
hierarchy of strategies used to find a proof.  Of course, there are
choices which hierarchy-level to consider and which form of hint
(Socratic vs. didactic, next step vs. strategic) to deliver in a
specific situation.  Following ideas from
\cite{DBLP:conf/its/BarnesS08} 
the tutor system itself could record
the hints given away in specific proof situations and try to assess
how useful they were.  From that historical tutoring data it could
come up with a classifier deciding which form of hint to use in which
situation.

\section{Related Work}
\label{sec:related work}

We discuss our work in connection with related 
approaches that (i) focus on domain reasoning techniques 
for tutoring proofs in logics and mathematics (\APROS and the Carnegie 
Proof Lab \cite{DBLP:journals/igpl/Sieg07}, the EPGY 
Proving Environment \cite{SommerNuckols04}, 
Tutch \cite{ACP-01-a} and approaches using hierarchical 
proofs), and (ii) hint generation (Carnegie Proof Lab, \ANDES\cite{Gertner98proceduralhelp} and 
the NovaNet Proof Tutorial~\cite{BarnesAndStamper2010}).\ednote{MS: 
shall we include Dimitra's thesis as related work?\\AS@MS: Yes, please}


\paragraph{APROS.} 
The \APROS project (see \cite{DBLP:journals/igpl/Sieg07} 
for an overview) provides an
integrated environment for strategic proof search and tutoring in
natural deduction calculi for predicate logic. It consists of four
modules: the \emph{proof generator} which implements strategic proof
search based on the intercalation calculus, the \emph{proof lab},
which builds the interface to the students, the \emph{proof tutor},
which generates hints for students that are stuck, and a web-based
course containing additional learning material. Proofs are represented
in a Fitch-style diagram and constructed by adding/removing steps to
the diagram. This means that the student enters stepwise the proof at
the calculus level, and that the performed steps can therefore
immediately be checked. If a student requests a hint, the proof
generator initiates the construction of a complete proof, which the
tutor analyzes to extract a hint. The first hint provided at any point
in the proof is a general strategic one, and subsequent hints provide
more concrete advice as to how to proceed. The last hint in the
sequence recommends that the student take a particular step in the
proof construction.

Compared with \APROS, which focuses on the teaching of one particular
logical calculus (without equality), the main difference with our
approach is that we focus on the teaching of more abstract assertion
level proofs, which are rather independent of a particular logical
calculus. Proofs are essentially constructed in a declarative proof
language in the form of proof sketches. If the information provided by
the student is complete and correct, the verification is just a simple
checking, as in the case of \APROS. However, within our setting, this
is not the typical situation.  Rather, it is common that the
information provided by the student is incomplete, as humans typically
omit information they consider unimportant or trivial. Therefore,
reconstruction of the missing information is necessary, as well as an
analysis of the complexity of this information.

The generation of hints is similar to our approach in the sense that
(i) it is dynamic and based on a completion of the student's proof
attempt, and (ii) that it can be provided at several levels of
granularity. However, we do not focus on a particular calculus,
neither on a fixed proof strategy. This necessitates the sophisticated 
techniques presented in Section~\ref{sec:proof step analysis} to instantiate the system 
with the required problem solving knowledge in the form of assertions 
and proof strategies, and to model the student's knowledge via a student 
model, which is used for granularity analysis.


\paragraph{EPGY.} 
The \EPGY theorem proving environment aims to support
``\emph{standard mathematical practice}'' both in how the final proofs
look as well as the techniques students use to produce them (see
\cite{SommerNuckols04} p.~227). To verify the proof steps entered by a
student, \EPGY relies on the CAS \MAPLE and the ATP \OTTER. The system
is domain independent in the sense that the course authors can specify
the theory in which a particular proof exercise takes place. Proof
construction works by selecting predefined rules and strategies from a
menu, such as definition expansion or proof by contradiction, or by
entering formulas. For computational transformations, a so-called
derivation system is provided. Once a statement is entered, the
student selects a set of justifications that he thinks is sufficient
to verify the new statement. The assumptions together with the goal
and implicit hidden assumptions are then sent to \OTTER with a time
limit of four to five seconds to verify the proof step.

Compared to our approach, the main similarities are that the system
aims at teaching ordinary mathematical practice independent of a
particular calculus. Moreover, it uses a theorem prover as domain
reasoner to dynamically verify statements entered by a student. The
authors acknowledge that the use of a classical ATP to verify proof
steps has the following drawbacks (\cite{SommerNuckols04} p.~253-254):
(i) ``One weakness of the Theorem Proving Environment is that, like
most computer-based learning tools, it does not easily assess the
elegance and efficiency of the student's work''. (ii) ``In the current
version of the Theorem Proving Environment, students are not given any
information as to why an inference has been rejected. Students are
told generally that an inference may be rejected because it represents
too big a step of logic, because the justifications are insufficient
to imply the goal, or because the goal is simply unverifiable in the
current setting. From our standpoint, Otter's output is typically not
enough to decide which is the reason of failure''.

In contrast, our approach relies on using the assertion level as a
basis to verify statements uttered by a student. This results in an
abstract proof object, which can further be analyzed, for example with
respect to granularity, as demonstrated in 
Section~\ref{sec:proof step analysis}, or to
extract hints on how to proceed if the student gets stuck. In
particular, the problem whether specified assertions were used in the
derivation can trivially be solved. We believe that limiting the
runtime of \OTTER does not reveal any information about the complexity
of a particular proof step. While it would also be possible to analyze
the resulting proof object, we believe that it is not at an
appropriate level of granularity and does not reflect a human-style of
proof construction. Even for natural deduction calculi, an 
investigation~\cite{lpar2006} into the correspondence between human
proofs and their counterparts in natural deduction points out a
mismatch with respect to their granularity.

Moreover, our approach is more flexible with respect to the following
aspects: (i) Due to the use of a proof language, the student is more
flexible in entering the solution. (ii) It is compatible with the
buggy rule approach. Note that this is not the case for classical
automated reasoners, in which an inconsistent theory makes everything
provable. In contrast, our approach allows for a full control over
buggy rules, such as limiting their application to a single
step. (iii) Our approach supports incomplete information such as a
missing subgoal. Note that by leaving out such a subgoal, the
resulting proof obligation becomes unverifiable and can therefore not
be supported by a classical ATP. Finally, our approach is extensible
and supports the specification of domain-dependent proof strategies,
as well as checking whether a particular step can be checked by a
specified proof strategy. This is not possible in the work cited
above.

\paragraph{TUTCH.} 
\TUTCH (see \cite{ACP-01-a}) is a proof checker that was originally 
designed for
natural deduction proofs in propositional logic. However, it was later
extended to also feature constructive first order logic to support
human oriented proof steps. To that end, a simple proof language that
allows steps at the assertion level was developed, as well as proof
strategies that allow for an efficient proof checking for proofs
within that language. This is similar to our approach, which also
relies on a proof language as well as a dynamic reconstruction of the
proof steps. Because of these similarities, we focus on the details of
the proof language and the strategies to verify the proof steps.

\paragraph{Hierarchical Proofs.}
Hierarchical proofs have been advocated by several people, such as
Lamport~\cite{Lamport93howto} in the context of informal proofs. A
similar idea is proposed by Back and colleagues for calculational
proofs~\cite{Back96structuredcalculational,Back2009}. 
In the context of HOL,
Grundy and Langbacka~\cite{DBLP:conf/amast/GrundyL97} developed an
algorithm to present hierarchical proofs in a browsable format.
Another possibility for hierarchical
proof construction is provided by a method called \emph{window
  inference}~\cite{DBLP:journals/logcom/RobinsonS93}. Window inference allows the user to
focus on a particular sub-formula of the proof state, transforming it
and thereby making use of its context, as well as opening subwindows,
resulting in a hierarchical structure.

Our approach is based on previous work by Cheikhrouhou and Sorge 
who developed the hierarchical proof data structure PDS 
in an earlier version of the \OMEGA system, intended to support 
hierarchical proof presentation and proof search \cite{CheikhrouhouSorge00pds}.  
In particular, the PDS supports so-called ``island proofs'', 
i.e. proofs that contain gaps which can be filled via refinement 
operations. The same idea has been picked up by
Denney, who developed the notion of hiproof~\cite{hiproofs2006}. Most
recently, a tactic language for hiproofs has been proposed
in~\cite{AspinallDenneyLueth10}.

\paragraph{MathsTiles.}

MathsTiles~\cite{DBLP:journals/jar/BillingsleyR07} are a flexible language to be used as an interface for 
an intelligent book on mathematics. In particular, proof exercises are offered 
where proofs formulated by the student are checked via Isabelle/HOL.
 The tiles correspond to graphical elements 
that can be arranged and recombined within a mathematical document, 
and which are used to represent formulas. This allows for flexibility while 
writing the proofs (e.g.~in the order in which proof lines are written). However, 
the approach in \cite{DBLP:journals/jar/BillingsleyR07} checks proofs linearly. 
In this context, the question of proof granularity 
is discussed. The authors state that the students should not use the 
prover to solve the exercises for them, and therefore the student is limited to using 
only Isabelle's simplifier (simp). Since rules can be added or removed from the 
simplifier, this makes the approach configurable. 
In contrast to our approach, however,  such rule sets within the mathematical 
domain represent only an implicit model of granularity 
which does not take into account dynamic information such as the proof context 
and the student's knowledge.

\paragraph{\ANDES.} 

\ANDES~\cite{Gertner98proceduralhelp} is an ITS for teaching Newtonian
physics. Like in our approach, student and
 tutor solve problems collaboratively, 
which is called \emph{coached problem solving}. \ANDES 
includes a problem solver, which is run on the problem description 
to generate a solution graph. The system uses abstract plans, 
such that the resulting solution graph represents 
a hierarchical dependency network.   
Similar to our approach, \ANDES uses templates to generate hints based
on the solution graph \cite{Gertner98proceduralhelp}. There may be 
several paths to a solution. \ANDES uses a 
Bayesian network for plan recognition to determine on which 
solution path the student might be on for giving an appropriate hint. 
As a justification, the authors mention their own informal studies 
where they found that human tutors rarely ask students about their 
goals before giving a hint. This can also be considered a motivation 
for our approach, where we keep 
track of several possible proof reconstructions simultaneously. 
Similarly to our approach again, the hints that are generated by 
\ANDES range from general to specific.
  

\paragraph{NovaNet Proof Tutorial.}

The NovaNet Proof Tutorial~\cite{BarnesAndStamper2010} is a learning 
tool for logic proofs. Students write proofs line by line which are verified
by the system. To generate hints for the next step, a 
 path through the space of the previously explored actions is sought 
 by optimizing a Markov decision process. This substitutes the use 
 of (strategic) proof search by data-mining a large number of example
 solutions. \cbstart[4pt]
The process can be tuned to extract an expert, a typical, or least 
error-prone solution. 
For the suggested steps, four levels of hint are generated 
and presented to the student in sequence; (i) the goal is indicated, 
(ii) what rule is to be applied, (iii) the statement the 
rule can be applied to, (iv) both the rule and the statements
(bottom-out-hint). 
These variants of hints are a subset of the hinting categories provided by our approach as 
presented in Section \ref{sec:hinting}, and do not involve proof hierarchies.  
\cbend

\section{Conclusion}
\label{sec:conclusion} 
In this paper we presented a coherent overview of the design of and
methods for an ITS to teach students to write
mathematical proofs in textbook style. Based on a detailed analysis of
the teaching domain, the paper argues in favor of adopting the
model-tracing tutor style to support the inner loop of such an
ITS. The tutor was then built on top of the slightly adapted proof
assistant \OMEGA to provide step analysis and hint generation.
Feedback is provided on each proof step entered by the student in form
of a vector composed of the criteria soundness, relevance and
granularity.  Hints are provided with increasing degree if
explicitness.

The following features of \OMEGA were particularly important to
realize the system: First, \OMEGA's declarative proof script language
allowed to define a clean interface proof sketch language to separate
natural language analysis from the pure step analysis and hint
generation tasks. Second, the assertion-level proof calculus allowed
for a depth limited search to reconstruct missing information in proof 
steps. 
We established that information obtained by proof reconstruction 
is a good basis for classifier-based granularity analysis. In particular, 
this allows the system to judge granularity based on criteria 
such as the number of assertions used by the student 
(which was found to be a useful criterion for modelling human tutors' judgments),  
the number of which are mastered or unmastered by the student, etc. 
We furthermore investigated the use of machine learning techniques to 
instantiate the classification module via corpora of example classifications 
from human experts. 
Third, \OMEGA's strategy language combining
declarative and procedural LCF-style tactics served as an authoring
language for domain specific strategic proof procedures. They can be
used both to generate a proof from scratch as well as to complete the
partial proof of the student. Recording the hierarchy of strategies
completing a proof provided an excellent basis for generating hints. 

The resulting prototype tutoring system has been evaluated on a corpus
of tutorial dialogues to analyze the student inputs and yields good
results.  Having a proof assistant system as domain reasoner bears a
lot of potential in order to improve the quality of proof step
analysis, feedback, and reviewing of solutions.  Further work is also
devoted to design the interface towards the student, which maps the
student input to the intermediate formal proof step format.

\bibliographystyle{eptcs}
\bibliography{paper}



\end{document}

%% file: diagram.tex

\tikzstyle{component}=[rectangle, rounded corners, 
                                    thick,
                                    minimum height=0.8cm,
									minimum width=3.8cm,
                                    draw=purple!80,
                                    fill=purple!20]

\tikzstyle{connection} = [line width=0.03cm,single arrow head indent=.4ex]
\pgfdeclarelayer{background}
\pgfsetlayers{background,main}
\begin{tikzpicture}
 \matrix[row sep=1cm,column sep=1.8cm] {
        \node (nl) [component]{NL Analyzer};  \\
        \node (stepanalysis) [component]{Proof Step Analysis};  
    & \node (granularityanalysis) [component]{Granularity Analysis}; \\
        \node (hint) [component]{\begin{tabular}{c}Error Feedback \& \\Hint Generation\end{tabular}};  
     & & \node (feedback) [component]{Immediate Feedback}; \\
    }; 
   \path [connection, ->] (nl) edge node[right] {ok} (stepanalysis);
   \path [connection,->,  draw=black] (stepanalysis) edge node[above] {ok}  (granularityanalysis);
   \path [connection, ->, draw=black] (stepanalysis) edge node[right] {not ok}  (hint);
   \path [connection, ->, draw=black] (granularityanalysis) -| node[below right] {ok} (feedback);
   \path [connection, ->, draw=black] (granularityanalysis) |-  node[above left] {not ok} (hint);
\begin{pgfonlayer}{background}
\node[rectangle,fill=yellow!25,inner sep=12pt,dashed, fit=(stepanalysis) (granularityanalysis) (hint) (feedback)] (pacbox) {};
\node[rectangle,draw,inner sep=12pt, fit=(nl) (stepanalysis) (granularityanalysis) (hint) (feedback)] (pacbox) {};
\end{pgfonlayer}
\end{tikzpicture}